\documentclass[11pt]{article}

\usepackage[T1]{fontenc}
\usepackage[utf8]{inputenc}
\usepackage{hyperref}
\usepackage{url}
\usepackage{booktabs}
\usepackage{amsfonts}
\usepackage{nicefrac}
\usepackage{microtype}
\usepackage{wrapfig}
\usepackage{color}
\usepackage{authblk}
\usepackage{bm}

\usepackage[a4paper, total={6in, 8in}]{geometry}

\usepackage{todonotes}   

\usepackage{enumitem}
\usepackage{graphicx}
\usepackage{amsmath,amssymb}
\usepackage{amsthm}
\usepackage{bigstrut}
\usepackage{float}
\usepackage{caption}
\usepackage[ruled]{algorithm2e}
\usepackage{mathtools}

\newtheorem{theorem}{Theorem}
\newtheorem*{definition}{Definition}

\newtheorem{lemma}[theorem]{Lemma}
\newtheorem{cor}[theorem]{Corollary}

\newtheorem*{note}{Note}

\newtheorem{asmA}{Assumption}
\newtheorem{asmB}{Assumption}

\numberwithin{equation}{section}

\DeclareMathOperator*{\sgn}{sign}

\DeclareMathOperator*{\argmax}{argmax}
\usepackage{graphicx}
\usepackage{subcaption}

\newcommand{\vecb}{\bm{b}}
\newcommand{\vecx}{\bm{x}}
\newcommand{\vecv}{\bm{v}}
\newcommand{\vecw}{\bm{w}}
\newcommand{\vecz}{\bm{z}}

\newcommand{\PP}{\mathbb{P}}
\newcommand{\EE}{\mathbb{E}}

\newcommand{\R}{\mathbb{R}}
\newcommand{\real}{\mathbb{R}}

\newcommand{\mX}{\mathcal{X}}
\newcommand{\mY}{\mathcal{Y}}
\newcommand{\mH}{\mathcal{H}}
\newcommand{\mM}{\mathcal{M}}
\newcommand{\mR}{\mathcal{R}}
\newcommand{\mC}{\mathcal{C}}
\newcommand{\mF}{\mathcal{F}}
\newcommand{\mG}{\mathcal{G}}
\newcommand{\mP}{\mathcal{P}}
\newcommand{\mD}{\mathcal{D}}
\newcommand{\mT}{\mathcal{T}}
\newcommand{\mB}{\mathcal{B}}
\newcommand{\II}{\mathbb{I}}
\newcommand{\JJ}{\mathbb{J}}

\usepackage[round]{natbib}

\title{A Geometric Perspective on the Transferability of Adversarial Directions}

\author{Zachary Charles}
\author{Harrison Rosenberg}
\author{Dimitris Papailiopoulos}

\affil{\tt{zcharles@wisc.edu hrosenberg@ece.wisc.edu dimitris@papail.io}   }

\affil{University of Wisconsin--Madison}

\date{}

\begin{document}

\maketitle

\begin{abstract}
State-of-the-art machine learning models frequently misclassify inputs that have been perturbed in an adversarial manner. Adversarial perturbations generated for a given input and a specific classifier often seem to be effective on other inputs and even different classifiers. 
In other words, adversarial perturbations seem to transfer between different inputs, models, and even different neural network architectures. 
In this work, we show that in the context of linear classifiers and two-layer ReLU networks, there provably exist directions that give rise to adversarial perturbations for many classifiers and data points simultaneously.
We show that these ``transferable adversarial directions'' are guaranteed to exist for linear separators of a given set, and will exist with high probability for linear classifiers trained on independent sets drawn from the same distribution.
We extend our results to large classes of two-layer ReLU networks. We further show that adversarial directions for ReLU networks transfer to linear classifiers while the reverse need not hold, suggesting that adversarial perturbations for more complex models are more likely to transfer to other classifiers. 
We validate our findings empirically, even for deeper ReLU networks.
\end{abstract}


\section{Introduction}

Many popular machine learning models, including deep neural networks, have been shown to be vulnerable to adversarial attacks \citep{szegedy2013intriguing,goodfellow2014explaining, nguyen2015deep, moosavi2016deepfool}. Small adversarial perturbations of image data can cause the model to significantly misclassify the data, even though the perturbed image may seem unchanged to the human eye. These perturbations are highly structured, as neural networks have been shown to be robust to random perturbations \citep{liu2016delving,fawzi2016robustness}.
While there has been a significant amount of work on designing adversarial attacks \citep{grosse2016adversarial,moosavi2016deepfool,mopuri2017fast,hendrik2017universal, papernot2016transferability} and defenses against these attacks \citep{madry2017towards, sinha2017certifiable}, theoretical properties of these adversarial perturbations are not fully understood. As shown by \citet{athalye2018obfuscated}, state-of-the-art defenses are often beaten in short order by newly designed attacks. A better theoretical understanding of these adversarial examples could lead to a better understanding of why attack and defense strategies perform well in certain situations and badly in others.

One phenomenon that has been repeatedly observed in the literature is that adversarial perturbations often {\it transfer} to other data points. For instance, \citet{moosavi2017universal} show that adversarial perturbations for a given input often work as an adversarial direction for many other inputs on the same neural networks. Such adversarial perturbations are often referred to as ``universal'' adversarial perturbations. Even worse, adversarial perturbations often transfer between classifiers \citep{papernot2016transferability,liu2016delving}. This seems to hold even if the classifiers have different architectures and are trained on different subsets of the training data \citep{szegedy2013intriguing}.

This transferability has led to more effective adversarial attacks. \citet{narodytska2016simple} show that even if an adversary only has black-box access to a neural network, they can still fool it relatively often. \citet{papernot2016transferability} construct adversarial examples with only black-box access to a neural network by generating adversarial examples on a substitute network designed to emulate the first. Recent work has attempted to make machine learning systems more secure by ``blocking'' transferability \citep{hosseini2017blocking}, to only limited success.

Unfortunately, adversarial perturbations are not yet fully understood, especially on a theoretical level. \citet{moosavi2017analysis} show that adversarial perturbations exist if certain geometric conditions hold, while \citet{fawzi2018analysis} study the robustness of linear and quadratic classifiers to adversarial examples. \citet{tramer2017space} show empirically that subspaces corresponding to the span of adversarial examples often have large intersection, demonstrating the capacity of adversarial examples to transfer. Still, it is not fully understood why and which adversarial perturbations transfer between classifiers, even for relatively simple classifiers.

\paragraph{Our Contributions: }In this work, we show that for linear classifiers and certain two-layer ReLU networks, classifiers trained on similar data sets can often be targeted by similar adversarial perturbations. Moreover, the geometry of the decision boundary causes these perturbations to be adversarial for most of the training data. We do this by analyzing directions which lead to adversarial perturbations, so-called ``adversarial directions.'' While many adversarial directions do not transfer to other classifiers and data, we explicitly construct adversarial directions which will transfer with high probability.

In Section \ref{sec:single_svms}, we show that for linear classifiers, the max-margin SVM can be used to construct adversarial directions which transfer to other data points and to other linear classifiers. We also show that for all linear separators of a given dataset, there are universal adversarial perturbations (that is, perturbations which are adversarial for all points with a given label) whose norm depends only on the max-margin classifier. We extend our results to soft-margin linear classifiers trained on independent data sets, as well as multi-class linear classifiers. In Section \ref{sec:relus}, we consider adversarial perturbations for ReLU networks. We use a geometric analysis to show that there exist adversarial directions for certain two-layer ReLU networks which transfer to all other such ReLU networks and all the training data of a given class. In Section \ref{sec:svms_relus} we show that while these adversarial direction for ReLU networks transfer to linear classifiers, the reverse need not hold. In Section \ref{sec:experiments}, we augment our theory with an empirical study showing that adversarial perturbations often transfer within a classifier, and even between distinct classifiers.

\section{Overview}\label{sec:setup}

	\paragraph{Notation:} In the following we denote vectors in bold script, and functions, sets, and scalars in standard script. Matrices are denoted by capital letters. For a vector $\vecw$, $w_i$ denotes its $i$th element.

	Suppose we have some data space $\mX \subseteq \R^d$ with label space $\mY$. For any set $S \subseteq \mX \times \mY$ and $a, b \in \mY$, let $S[a] = \{(\vecx,y) \in S : y = a\}$ and $S[a,b] = \{(\vecx,y) \in S : y\in \{a,b\}\}$. A classifier is a function $h: \mX \to \mY$. We say that $h$ is $S$-accurate if for all $(\vecx,y) \in S$, $h(\vecx) = y$. We are interested in when we can perturb $\vecx \in \mX$ to produce $\vecx'$ such that $h(\vecx') \neq h(\vecx)$. This leads to the following definition.

	\begin{definition}We say that $\vecv$ is an adversarial direction for $h$ at $(\vecx,y)$ if $h(\vecx) = y$ and there is some $c > 0$ such that $h(\vecx+c\vecv) \neq y$.\end{definition}
	The vector $c\vecv$ is referred to as an {\it adversarial perturbation}, while the vector $\vecx' = \vecx + c\vecv$ is referred to as an {\it adversarial example}. We will show that for well-known classes of classifiers, there are directions that are adversarial for many classifiers on large subsets of $\mX$. In other words, the adversarial direction is ``transferable'' both to other classifiers and to other data points.

\begin{figure}[ht]
		\centering
		\includegraphics[width=0.5\linewidth]{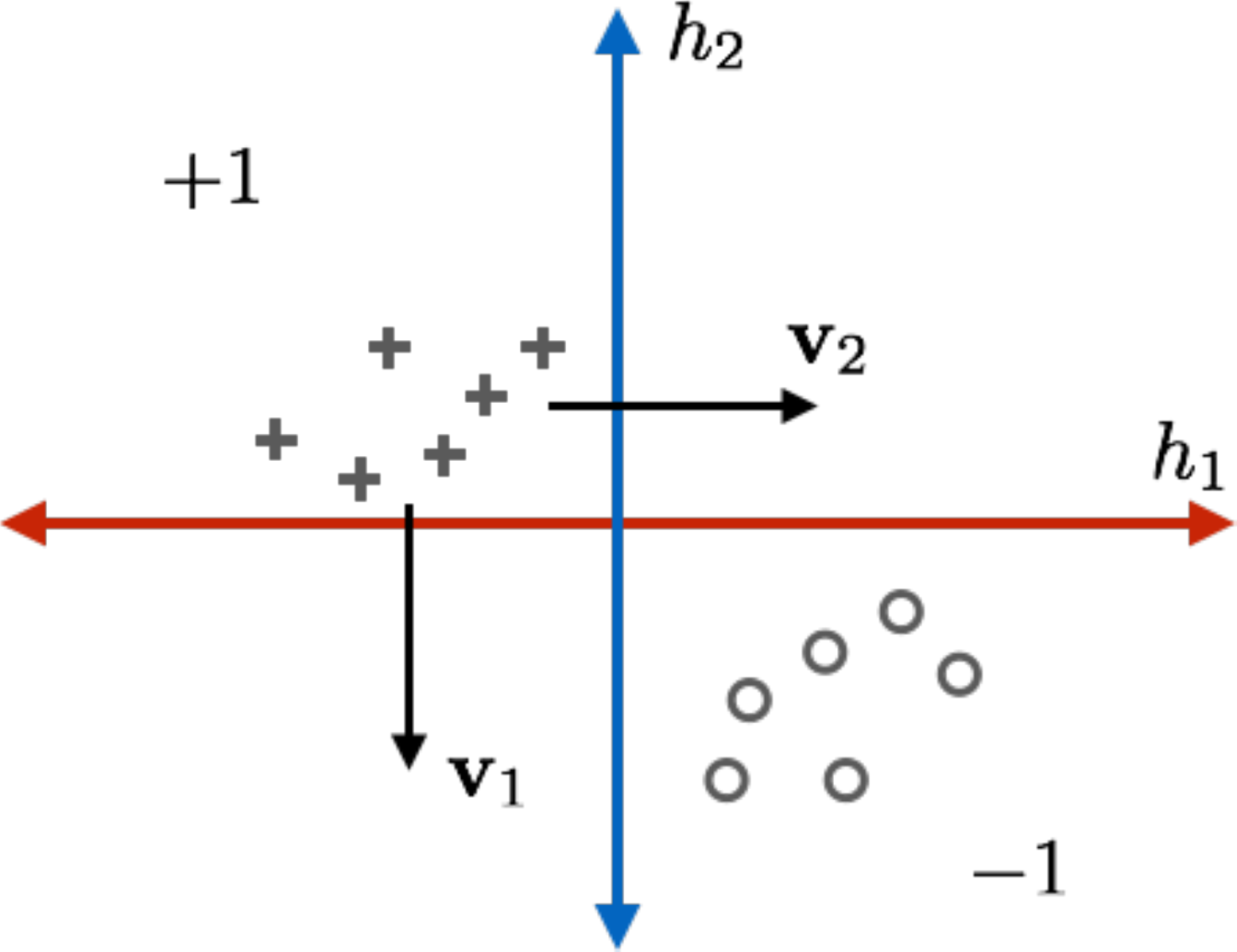}
		\caption{\small The decision boundary of two linear classifiers, $h_1$ (red) and $h_2$ (blue). The data with label $+1$ are marked by pluses, and those with label $-1$ are marked by circles. For the $+1$ data, $\vecv_1$ and $\vecv_2$ give adversarial directions for $h_1$ and $h_2$, respectively.}
		\label{fig:svms}
	\end{figure}

	The fact that adversarial directions transfer is not immediately obvious or even true for all models. For example, consider Figure \ref{fig:svms}. For any linear classifier, the most natural adversarial direction is the normal vector to the decision boundary. If we consider just the linear classifier $h_1$, this direction is given by $\vecv_1$. However, $\vecv_1$ is not an adversarial direction for $h_2$, as moving in that direction does not change any labels. Similarly, $\vecv_2$ is adversarial for $h_2$ but not for $h_1$. In short, not all adversarial directions transfer.

	We instead show that the {\it max-margin classifier} for a set $S$ gives us an adversarial direction which transfers to all classifiers that linearly separate $S$, and extend this to the case where we train two linear classifiers $h_1, h_2$ on $S_1, S_2$ sampled independently from some distribution $\mD$. We derive the following theorem.

	\begin{theorem}\label{thm:informal1}Suppose linear classifiers $h_1, h_2$ are trained on sets $S^{(1)}, S^{(2)}$ of size $n$ sampled independently from $\mD$, and that each $h_j$ correctly classifies $\Theta(n)$ of $S^{(j)}$ correctly. With high probability, there is a vector $\vecw$ and a set $S \subseteq S^{(1)} \cup S^{(2)}$ such that $\vecw$ is an adversarial direction for both $h_1$ and $h_2$, on $S$ and $|S\cap S^{(j)}| = \Theta(n)$ for $j = 1,2$.\end{theorem}

	This is an informal version of Theorem \ref{thm:soft_margin}. Intuitively, as long as we train linear classifiers on similar datasets, there is an adversarial direction which transfers between the linear classifiers and works for most of the data. This occurs even if $h_1, h_2$ are not the max-margin SVM on $S^{(1)}, S^{(2)}$. We derive a version of this for multi-class linear classifiers in Theorem \ref{thm:multi_soft_svm}.

\begin{figure}[ht]
		\centering
		\includegraphics[width=0.5\linewidth]{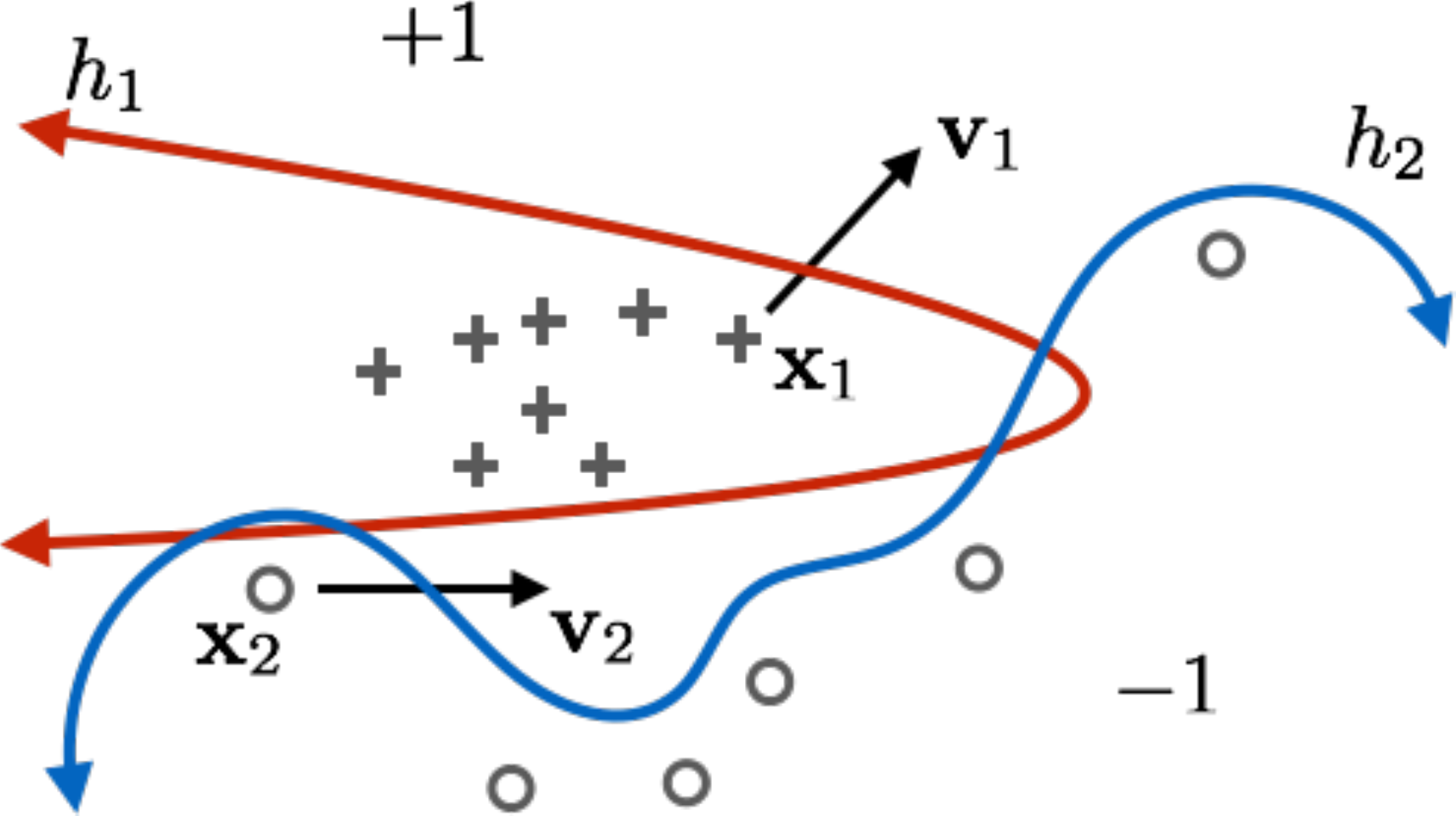}
		\caption{\small Two nonlinear decision boundaries, $h_1$ (red) and $h_2$ (blue). The instances with label $+1$ are marked by pluses, those with label $-1$ are marked by circles. An adversarial direction for $h_1$ at $\vecx_1$  is denoted by $\vecv_1$, while an adversarial direction for $h_2$ at $\vecx_2$ is denoted by $\vecv_2$.}
		\label{fig:nns}
	\end{figure}

	For neural networks, the study of adversarial directions becomes more difficult. For a linear classifier, an adversarial direction for some data point is also adversarial for other data points with the same label. This need not hold for neural networks. Consider Figure \ref{fig:nns}. There are two nonlinear decision boundaries for classifiers $h_1, h_2$. Note that $\vecv_1$ is adversarial for $h_1$ at $\vecx_1$ and for the other points with $+1$ label. However, $\vecv_1$ does not transfer to the $+1$ instances for $h_2$. Furthermore, $\vecv_2$ is an adversarial direction for $h_2$ at $\vecx_2$, but does not transfer to all other points with label $-1$ for $h_2$, nor does it transfer to any $-1$ labeled point for $h_1$.

	With arbitrarily complicated decision boundaries, the study of adversarial directions becomes difficult. If we impose structure on our neural networks, we can begin to understand them. We show that for large classes of two-layer ReLU networks, there are transferable adversarial directions. The following theorem is an informal version of Theorem \ref{thm:relu_direction}.

	\begin{theorem}\label{thm:informal2}Fix a set $S$ and consider the set of two-layer ReLU network with nonnegative weights on the final layer that correctly classify $S$. There is a vector $\vecv_S$ that is adversarial for all such networks at any point $(\vecx,y) \in S$ such that $y = -1$.\end{theorem}

	We also study whether adversarial directions transfer between linear classifiers and ReLUs. In Section \ref{sec:svms_relus} we show that while the adversarial direction for ReLU networks in Theorem \ref{thm:informal2} transfers to linear classifiers, the adversarial direction for linear classifiers in Theorem \ref{thm:informal1} does not necessarily transfer to ReLUs. In general, it seems that adversarial directions are more likely to transfer to a classifier with a less complicated decision boundary than with a more complicated one. We give supporting empirical evidence for this phenomenon in Section \ref{sec:experiments}.

\section{Linear Classifiers}\label{sec:single_svms}

	\subsection{Hard-margin Classifiers}

	We first focus on the case of hard-margin linear classifiers trained on the same training set $S$. By hard-margin, we mean that the classifier should correctly classify every point in $S$. We will show that the max-margin SVM classifier on $S$ gives a transferable adversarial perturbation between the linear classifiers. We will relate the size of the perturbation needed to the support vector coefficients. We will use this as a stepping stone to consider both soft-margin linear classifiers and training data sampled from some underlying distribution.

	Suppose we have data $\mX \subseteq \real^d$ with labels $\{\pm 1\}$. An linear classifier $h$ is given by $h(\vecx) = \sgn(\langle \vecw,\vecx\rangle + b)$ for some $\vecw \in \R^d, b \in \R$. Note that $h(\vecx)$ is invariant under scaling $(\vecw, b)$ by the same positive constant, so we restrict to $\|\vecw\|_2 = 1$. We let $\mH$ denote the set of such linear classifiers.

	Suppose we have some set $S$ of labeled examples $\{(\vecx,y)\}_{i=1}^N$ and that $h$ is $S$-accurate. We will let $\mH_S := \{ h \in \mH \mid h\text{ is }S\text{-accurate}\}$. The {\it margin} of $h \in \mH_S$ is defined as $\gamma(h) = \min_{(\vecx,y) \in S} y(\langle \vecw,\vecx\rangle+b)$.

	We say that $S$ is {\it linearly separable} if $\mH_S \neq \emptyset$. In this case, standard theory shows that there is an SVM classifier $h^* \in \mH_S$ with maximum margin given by $h^* = \sgn(\langle \vecw^*,\vecx\rangle + b^*)$. We will let $\gamma^*$ denote its margin. We will use the following lemma about max-margin classifiers on $S$. This follows from Theorem 15.8 of \citet{shalev2014understanding} and the discussion thereafter, combined with strong duality.
	\begin{lemma}\label{lem:max_marg_w}There are $\alpha_i \geq 0$ such that (1) $w^* = \sum_{i=1}^N \alpha_iy_i\vecx_i$ and (2) $\sum_{i=1}^N \alpha_iy_i = 0$.
	\end{lemma}
	We will use this to show that the max-margin classifier gives us a transferable adversarial direction.
	\begin{theorem}\label{thm:svm_adv_dir}For all $h \in \mH_S$ and $(\vecx,y) \in S$, $-y\vecw^*$ is an adversarial direction for $h$ at $\vecx$.\end{theorem}
	A slight modification of the proof shows the following.
	\begin{theorem}\label{thm:svm_adv_dir2}Fix $h \in \mH_S$ with margin $\gamma_h$. Fix $(\vecx,y)$ such that $y(\langle\vecw,\vecx\rangle+b) > 0$. Then $h(\vecx-cy\vecw^*) \neq h(x)$ for all $c$ satisfying
	$$c > \dfrac{y(\langle\vecw,\vecx\rangle+b)}{\gamma_h \sum_{i=1}^N \alpha_i}.$$\end{theorem}
	
	Fix a linear classifier $h$. If we take $(\vecx,y)$ to be the point in $S$ with minimum margin for $h$, this implies that as long as $c > (\sum_{i=1}^N\alpha_i)^{-1}$, we will have $h(\vecx-cy\vecw^*) \neq h(\vecx)$. Thus, for this $c$, we are guaranteed that perturbing the data by $cy\vecw^*$ will cause $h$ to misclassify some $\vecx$. This implies the following corollary.
	\begin{cor}For any $h \in \mH_S$, there is some $(\vecx,y) \in S$ such that for all $c > (\sum_{i=1}^N \alpha_i)^{-1}$, $h(\vecx-cy\vecw^*) \neq h(\vecx)$.\end{cor}
	Thus, for any set $S$, there is a universal constant $c_S$ such that for any $h$ that linearly separates $S$ and any $(\vecx,y) \in S$, $-c_Sy\vecw^*$ is an adversarial perturbation for $h$ at $\vecx$.

	\subsection{Soft-margin Classifiers}

	The results above have two fundamental limitations. First, they only concern hard-margin linear classifiers. Second, they assume that the classifiers are trained on the same dataset $S$. Here, we extend the results above to soft-margin linear classifiers trained on samples drawn independently from some distribution. We will show that if we train two separate soft-margin linear classifiers on independently drawn training sets, then with high probability, there will be a single adversarial direction for both classifiers that affects most of the training examples in both sets.

	Let $\mD$ be a distribution over $\mX\times \mY \subseteq \real^d \times \{\pm 1\}$. We assume that with probability $1$ over $\mD$, $\|\vecx\|_2 \leq R$. Let $S^{(1)}, S^{(2)}$ be training sets of size $n$ drawn independently from $\mD$. Suppose we have two linear classifiers $\vecw^{(1)}, \vecw^{(2)}$ trained on $S^{(1)}$ and $S^{(2)}$ in some way. For notational simplicity, we assume no bias terms, though the same result can be derived in this setting. Thus, the label assigned to $\vecx \in \mX$ by $\vecw^{(j)}$ is $h_j(\vecx) = \sgn(\langle\vecw^{(j)},\vecx\rangle)$. We make no assumptions about the training process itself. We only make three relatively minor assumptions on the output of the training process.

		\begin{asmA}\label{asm:A1}
		With probability $1$ over the training procedure, $\|\vecw^{(j)}\|_2 \leq B$.\end{asmA}

		\begin{asmA}\label{asm:A2}
		The probability that $\vecw^{(1)}$ correctly classifies any point in $S^{(1)}$ is independent from the probability $\vecw^{(2)}$ correctly classifies any point in $S^{(2)}$, and vice-versa.\end{asmA}

		\begin{asmA}\label{asm:A3}
		There are at least $(1-\rho)n$ for $\rho \in [0,1)$ points $(\vecx,y) \in S^{(j)}$ such that $y\langle \vecw^{(j)},\vecx\rangle\geq 1$.\end{asmA}

		\textbf{\ref{asm:A1}} is a straightforward assumption that states that the output of the training method cannot be of unbounded size, while \textbf{\ref{asm:A2}} essentially states that the training procedures are independent. \textbf{\ref{asm:A3}} simply says that $h_j$ is correct and {\it confident} about its prediction on at least a $(1-\rho)$ fraction of the dataset. Under these assumptions, we have the following theorem about adversarial directions for $h_1$ and $h_2$.

		\begin{theorem}\label{thm:soft_margin}Under the assumptions above, with probability at least $1-4n^{-2}$, there is a set $S \subseteq S^{(1)} \cup S^{(2)}$ with max-margin SVM $\vecw_S^*$ such that for $j \in \{1,2\}$,
		\begin{enumerate}
			\item $|S\cap S^{(j)}| \geq (1-2\rho)n - 2BR\sqrt{n} - 4\sqrt{n\ln(n)}$.
			\item For all $(\vecx,y) \in S$, $-y\vecw_S^*$ is an adversarial direction for $h_j$ at $\vecx$.
		\end{enumerate}
		\end{theorem}

		Note that for $\rho < \frac{1}{2}$ and $BR = o(\sqrt{n})$, $|S\cap S^{(j)}| = \Theta(n)$. Thus, there is a large subset of $S^{(1)} \cup S^{(2)}$ that both $h_1$ and $h_2$ are accurate on, but for which there is an adversarial direction $\vecv$ that transfers among all points in $S$ and both classifiers.

		To prove this, we first use Rademacher complexity bounds to show there is a large subset $S \subseteq S^{(1)}\cup S^{(2)}$ that both $h_1$ and $h_2$ classify correctly. Once we know that such an $S$ exists, we can use our hard-margin results to find an adversarial direction $\vecv$ that is adversarial for both classifiers on most of the union of the training sets. The details are in Appendix \ref{append:soft_margin}.

\section{Multi-class Linear Classifiers}\label{sec:multi_svms}

	\subsection{Hard-margin Classifiers}

	Suppose now that we have data $\mX \subseteq \real^d$ with $K$ potential labels. We will assume these labels are $\{1,2,\ldots, K\}$. We now consider multi-class classification. One standard way to extend linear classifiers to multi-class classification is to use a {\it one-versus-rest} approach. With this approach, we train $K$ binary linear classifiers. We train the $k$-th linear classifier on a version of $S$ with modified labels, where labels of $k$ are replaced with $+1$ and the remaining labels are replaced with $-1$. To classify $\vecx$, we evaluate it on all $K$ linear classifiers and return the class maximizing its output value.

	Formally, we consider classifiers of the form $h(\vecx) = g(W\vecx + \vecb)$ where $W \in \real^{K\times d}, \vecb \in \real^K$ and $g(\vecz) = \argmax_{1 \leq i \leq K}z_i$. Intuitively, $g(\vecz)$ returns the index $i$ of the largest entry of $\vecz$. We let $\mM$ denote the set of such classifiers.

	This setup generalizes standard techniques such soft-max layers. The soft-max function $s: \real^d \to \real^d$ turns any vector in to a probability vector, but satisfies $g(s(\vecz)) = g(\vecz)$. Therefore, our setup includes the soft-max function and any classifier $h$ that returns the index maximizing the vector $W\vecx+\vecb$. Let $S$ denote some set of instances with labels in $\{1,\ldots, K\}$. We will let $\mM_S$ denote the set of $S$-accurate classifiers.

	\begin{note}A linear classifier is $S$-accurate if $\langle \vecw,\vecx\rangle + b$ is positive for all positive instances and negative for the rest. For the multi-class case, $S$-accuracy does not require positivity of any classifier. Suppose $\vecx$ has label $k$, and let $\vecz = W\vecx+\vecb$. To classify $\vecx$ accurately we do not require $z_j < 0$ for $j \neq k$. In fact, all of $\vecz$ can be positive as long as the largest entry is $z_k$.\end{note}

	Recall that for $k \in \{1, \ldots, K\}$, we define $S[k] = \{(\vecx,y) \in S~|~y = k\}$. We will use max-margin classifiers to construct transferable adversarial directions for $S[k]$, as in the following theorem.

	\begin{theorem}\label{thm:multi_hard_svm}Fix $k \in \{1,\ldots, K\}$. There is a vector $\vecv_k^*$ such that for all $h \in \mM_S$ and for all $(\vecx,y) \in S[k]$, $\vecv_k^*$ is an adversarial direction for $h$ at $\vecx$.\end{theorem}

	A proof is given in the appendix. In fact, the theorem actually shows a slightly stronger statement. Namely, that for distinct $k, l \in \{1,\ldots, K\}$, there is a vector $\vecv_{k,l}^*$ that is an adversarial direction for all $h \in \mM_S$ and for all $(\vecx,y) \in S[k,l]$, where $S[k,l]$ was defined as $\{(\vecx,y) \in S~|~y \in \{k,l\}\}$.

	Note that it is not necessarily true that for sufficiently large $c$, for any $\vecx \in S[k]$, $h(\vecx-c\vecw^*) = m$. The direction $-\vecw^*$ may also increase the output associated to some other label. Since we know that $(W(\vecx-c\vecw^*)+\vecb)_k < (W(\vecx-c\vecw^*)+\vecb)_l$ for sufficiently large $c$, this is sufficient to show that $h(\vecx-c\vecw^*) \neq k = h(\vecx)$.

	\subsection{Soft-margin Classifiers}

	As in Section \ref{sec:single_svms}, we can extend our results to soft-margin classifiers trained on independently drawn sets. Suppose that we have some distribution $\mD$ on $\mX \times \{1,\ldots, K\}$ such that with probability $1$ over $\mD$, $\|\vecx\|_2 \leq R$. Let $S^{(1)}, S^{(2)}$ be training sets drawn independently from $\mD$. We have two models $W^{(1)}, W^{(2)}$ trained on $S^{(1)}$ and $S^{(2)}$ in some way again with no bias terms. Let $\vecw^{(j)}_i$ denote the transpose of the $i$-th row of $W^{(j)}$. The label assigned to $\vecx \in \mX$ by $W^{(j)}$ is given by $h_j(\vecx) = \max_{1 \leq i \leq K}\langle \vecw^{(j)}_i,\vecx\rangle.$
		
	As in Section \ref{sec:single_svms}, we do not make any explicit assumptions on the training process, only a few minor assumptions on the output models. Fix some $k \in \{1,\ldots, K\}$. Recall that for any set $S \subseteq \real^d \times \{1,\ldots, K\}$, $S[k]$ denotes the subset of $S$ with label $k$ and $S[k,l]$ denotes the subset with label $k$ or $l$. We assume the following three properties hold for $j \in \{1,2\}$.

		\begin{asmB}\label{asm:B1}
		With probability 1 over the training procedure, for $i \in \{1,\ldots, K\}$, $\|\vecw^{(j)}_i\|_2 \leq B$.\end{asmB}

		\begin{asmB}\label{asm:B2}
		The probability that $W^{(1)}$ correctly classifies any point in $S^{(1)}$ is independent from $S^{(2)}$ and $W^{(2)}$, and vice-versa.\end{asmB}

		\begin{asmB}\label{asm:B3}
		There is a label $l$ such that at least a $(1-\rho)$ fraction of points $(\vecx,y) \in S^{(j)}[k,l]$ satisfy $h_j(\vecx) = y$ and $|\langle \vecw^{(j)}_k,\vecx \rangle - \langle \vecw^{(j)}_l,\vecx\rangle| > 1$.\end{asmB}

		\textbf{\ref{asm:B1}} and \textbf{\ref{asm:B2}} are analogues of \textbf{\ref{asm:A1}} and \textbf{\ref{asm:A2}} in the linear classifier case and are relatively straightforward assumptions. \textbf{\ref{asm:B3}} is a kind of multi-class generalization of \textbf{\ref{asm:A3}}. It says that our classifier is approximately accurate on a fraction of samples with label $k$ or $l$, and moreover, that for such samples, $W^{(j)}$ is much more confident about the label $k$ than the label $l$. In other words, $W^{(j)}$ does not conflate the labels $k$ and $l$.

		For simplicity we assume that $|S^{(1)}[k,l]| = |S^{(2)}[k,l]| = n$. In general, we can derive an analogous result to Theorem \ref{thm:multi_soft_svm} below based on the ratio of $|S^{(1)}[k,l]|$ to $|S^{(2)}[k,l]|$. Note that as we sample more from $\mD$, with high probability $|S^{(1)}[k,l]|/|S^{(2)}[k,l]|$ will be close to 1. We then have the following theorem, whose proof we leave to the appendix.

		\begin{theorem}\label{thm:multi_soft_svm}Fix $k \in \{1,\ldots, K\}$ satisfying the above assumptions. With probability at least $1-4n^{-2}$, there is a vector $\vecv$ and a set $S \subseteq S^{(1)}[k,l] \cup S^{(2)}[k,l]$ such that for $j \in \{1,2\}$,
		\begin{enumerate}
			\item $|S\cap S^{(j)}[k,l]| \geq (1-2\rho)n - 4BR\sqrt{n} - 4\sqrt{n\ln(n)}$.
			\item For all $(\vecx,y) \in S$ such that $h_j(\vecx) = y$, $-\vecv$ is an adversarial direction for $h_j$ if $y = k$, and $\vecv$ is an adversarial direction for $h_j$ if $y = l$.
		\end{enumerate}
		\end{theorem}

		\begin{note}Condition 2 of Theorem \ref{thm:multi_soft_svm} is slightly weaker than condition 2 of Theorem \ref{thm:soft_margin}. While the adversarial direction still transfers, we do not require $h_1$ to correctly classify $S\cap S^{(2)}[k,l]$. This slight weakening comes about by reducing the multi-class linear classification case to the single-class linear classification case.\end{note}


\section{Two-layer ReLU Networks}\label{sec:relus}

	In this section we show that for certain classes of two-layer ReLU networks and any data set $S$, there is an adversarial direction for all networks in this class that separate $S$. In other words, if we restrict to certain classes of two-layer ReLU networks, then fitting to a data set $S$ forces the network to be susceptible in specific directions.

	Let $\sigma$ denote the ReLU function. For $z \in \real$, $\sigma(z) = \max\{0,z\}$. For $\vecz \in \real^n$, we abuse notation to let $\sigma(\vecz)$ denote the vector with $\sigma(\vecz)_i = \sigma(z_i)$. A two-layer {\it ReLU network} with hidden layer of width $L$ is a function $f: \real^d \to \real^k$ of the form $f(x) = V\sigma(W\vecx+\vecb)$.

	For simplicity, we will only consider binary classification with labels $\{\pm 1\}$. Therefore, we restrict to the setting where $k = 1$, so that there is only one output unit. In this case, a ReLU network is of the form $f(x) = \vecv^T\sigma(W\vecx+\vecb)$ for some $\vecv \in \real^L$. A {\it ReLU classifier} is a function $h:\real^d \to \{\pm 1\}$ of the form $h(\vecx) = g(f(\vecx))$ where $f$ is a ReLU network and $g: \real \to \{\pm 1\}$ is some decision rule. We will often consider $g$ that are indicator functions of the form $g(z) = 1$ for $z \in A$ and $-1$ otherwise. We denote such a function by $\JJ_A(z)$.

	We first consider ReLU networks $\mF$ where $\vecv$ is the all-ones vector and ReLU classifiers $\mR$ consisting of ReLU networks in $\mF$ combined with the classification rule $g$ where $g(x) = 1$ if $x > 0$ and $-1$ otherwise. Formally, we define:
	$$\mF := \left\{f: \real^d \to \real ~\middle\vert~ f(\vecx) = \sum_{i=1}^L \sigma(\vecw_i^T\vecx + b_i)\right\}.$$
	\begin{align*}
	\mR := \left\{h: \real^d \to \{\pm 1\} ~\middle\vert \begin{array}{l}
	h(\vecx) = g(f(\vecx)),\\
	g(z) = \mathbb{J}_{(0,\infty)}(z)
	\end{array}\right\}.\end{align*}
	Intuitively, such ReLU classifiers compute $L$ functions of the form $\sigma(\vecw_i^T\vecx+b_i)$ and then return 1 if at least one of these is positive.

	Let $S \subseteq \real^d \times \{\pm 1\}$ denote a set of labeled data. We assume that $S$ contains at least one example with each label. We let $\mR_S$ denote the set of $S$-accurate classifiers in $\mR$. We then have the following theorem that shows the existence of transferable adversarial directions for such classifiers.

	\begin{theorem}There is some $\vecv$ such that for all $h \in \mR_S$ and $(\vecx,-1)$ such that $h(\vecx) = -1$, $\vecv$ is an adversarial direction for $h$ at $\vecx$.\end{theorem}

	Note that the theorem implies more than $\vecv$ being adversarial for $h$ on $S$. In particular, it applies to any $(\vecx,-1)$ such that $h(\vecx) = -1$. Thus, $\vecv$ may be adversarial for more than just instances in the training set.

	We will derive this theorem as a consequence of a more general theorem. Suppose we have two possible labels, $\alpha$ and $\beta$. Consider all functions $h : \real^d \to \{\alpha,\beta\}$. Suppose we have some set $S \subseteq \real^d \times \{\alpha,\beta\}$. We will analyze $h$ such that $h^{-1}(\alpha)$ is convex (note that $h^{-1}(\beta)$ is not necessarily convex). We define
	$$\mC := \{h ~|~  h\text{ is continuous, } h^{-1}(\alpha)\text{ is convex}\}.$$
	We denote the set of $S$-accurate $h$ in $\mC$ by $\mC_S$. We then have the following theorem, which we prove in the appendix.

	\begin{theorem}\label{thm:convex_adv}There is some $\vecv$ such that for all $h \in \mC_S$ and $(\vecx,\alpha)$ such that $h(\vecx) = \alpha$, $\vecv$ is an adversarial direction for $h$ at $\vecx$.\end{theorem}

	The requirement that $h \in \mC_S$ can be weakened. Fix $\vecx_\alpha \in S[\alpha]$ and $\vecx_\beta \in S[\beta]$. Let $\mT$ denote the set of continuous functions $h: \real^d \to \{\alpha,\beta\}$ such that $h(\vecx_\alpha) = \alpha, h(\vecx_\beta) = \beta$, and there is some convex set $C_h$ containing $\vecx_\alpha$ such that for all $\vecx \in C_h^c$, $h(\vecx) = \beta$. A similar proof then shows the following result.

	\begin{theorem}Fix $\vecx_\alpha \in S[\alpha], \vecx_\beta \in S[\beta]$. There is some $\vecv \in \real^d$ such that for all $h \in \mT$ and for all $(\vecx,\alpha) \in C_h$ such that $h(\vecx) = \alpha$, $\vecv$ is an adversarial direction for $h$ at $\vecx$.\end{theorem}

	To apply Theorem \ref{thm:convex_adv} to $\mR_S$, it suffices to show the following lemma. We do so in the appendix.

	\begin{lemma}\label{lem:convex_class}For $h \in \mR$, $h^{-1}(-1)$ is convex.\end{lemma}

	We can extend $\mR$ to include a broader class of ReLU classifiers. Define
	$$\mF' := \left\{ f: \real^d \to \real ~\middle\vert \begin{array}{l}
	f(\vecx) = \vecv^T\sigma(W\vecx+\vecb),\\
	v_i \geq 0,~\forall i
	\end{array}\right\}.$$
	$$\mG := \left\{ g:\real\to \{\pm 1\} ~\middle\vert \begin{array}{l}
	g(z) = \JJ_{(a,\infty)}(z),~a > 0\\
	\text{or }g(z) = \JJ_{[a,\infty)}(z),~a > 0\end{array}
	\right\}.$$
	We can view $\mF'$ as the set of two-layer ReLU neural networks whose output layer has all nonnegative weights. We then define
	$$\mR' := \left\{ h:\real^d \to \{\pm 1\} ~\middle\vert \begin{array}{l}
	h(\vecx) = g(f(\vecx)),\\
	f \in \mF',~g \in \mG
	\end{array}
	\right\}.$$

	In order to apply Theorem \ref{thm:convex_adv}, we would need to show that one of the decision regions of any $h \in \mR'$ is convex. More formally, we have the following lemma.

	\begin{lemma}\label{lem:mr1_conv}
	For $h \in \mR'$, $h^{-1}(-1)$ is convex.\end{lemma}

	A proof is given in the appendix. Combining Theorem \ref{thm:convex_adv} and Lemma \ref{lem:mr1_conv}, we derive the following.

	\begin{theorem}\label{thm:relu_direction}There is some $\vecv$ such that for all $h \in \mR'_S$ and $(\vecx,-1)$ such that $h(\vecx) = -1$, $\vecv$ is an adversarial direction for $h$ at $\vecx$.\end{theorem}

	We can also extend this result to a class of ReLU classifiers $\mR''$ that are similar to $\mR$. Here, we consider the set of ReLU networks
	$$\mF'' := \left\{f: \real^d \to \real^L ~\middle|~ f(\vecx) = \sigma(W\vecx+\vecb)\right\}$$
	where $W \in \real^{L\times d}$ so that the output is potentially vector-valued and $\sigma$ is the ReLU function applied entry-wise. Let $\phi: \real^L \to \real$ where $\phi(\vecz) = 1$ if all entries of $\vecz$ are positive and $-1$ otherwise. We then define
	$$\mR'':= \left\{h:\real^d \to \{\pm 1\} ~\middle\vert\begin{array}{l}
	h(\vecx) =  \phi(f(\vecx)),\\
	f \in \mF''\end{array}\right\}$$
	Since $\phi(\vecz) = 1$ if all the entries of $\vecx$ are positive and $-1$ otherwise, $h^{-1}(1)$ is the intersection of $L$ open half-spaces corresponding to the pair $(W,\vecb)$. Therefore, $h^{-1}(1)$ is convex. Applying Theorem \ref{thm:convex_adv}, we then get the following theorem concerning the set $\mR''_S$ of $S$-accurate classifiers in $\mR''$.

	\begin{theorem}There is some $\vecv$ such that for all $h \in \mR''_S$ and $(\vecx,1)$ such that $h(\vecx) = 1$, $\vecv$ is an adversarial direction for $h$ at $\vecx$.\end{theorem}

\section{Transferability Between Linear Classifiers and ReLU Networks}\label{sec:svms_relus}

	In this section we analyze whether the adversarial directions above transfer between classifiers of different types. We will determine if and when adversarial directions for linear classifiers in Section \ref{sec:single_svms} transfer to the ReLU classifiers in Section \ref{sec:relus} and vice-versa.

	Suppose we have some training set $S \subseteq \real^d \times\{\pm 1\}$. We wish to consider applying both linear classifiers and ReLU classifiers to $S$. Recall that above, we defined $\mH$ to be the set of linear classifiers on $\real^d$. We also defined $\mR'$ as the set of two-layer ReLU classifiers of the form $h(\vecx) = g(f(\vecx))$
	where $g(z) = \JJ_{[a,\infty)}(z)\text{ or }g(z) = \JJ_{(a,\infty)}(z)$ and $f$ is of the form $f(\vecx) = \vecv^T\sigma(W\vecx+\vecb)$ with $\vecv \geq 0$. Recall that for a set $A$, $\JJ_A(z)$ is $1$ if $z \in A$ and $-1$ otherwise.

	We will let $\mH_S, \mR'_S$ denote the set of $S$-accurate classifiers in each of these sets. In Section \ref{sec:single_svms}, we constructed adversarial directions that applied to all $h \in \mH_S$ and all $(\vecx,y) \in S$, while in Section \ref{sec:relus}, we constructed adversarial directions for all $h \in \mR'_S$ and $(\vecx,y) \in S[-1]$, that is, examples with label $-1$. We will show that the adversarial direction for $\mR'_S$ constructed in Theorem \ref{thm:relu_direction} transfers to $\mH_S$, while the adversarial direction for $\mH_S$ constructed in Theorem \ref{thm:svm_adv_dir} need not transfer to $\mR'_S$.

	To see that the adversarial direction $\vecv$ for $\mR'_S$ transfers to $\mH_S$, recall that this direction was derived from Theorem \ref{thm:convex_adv}. In particular, all that is necessary for $\vecv$ to transfer to a classifier $h$ is that $h^{-1}(-1)$ is convex and that $\mH_S$ is non-empty. Since linear classifiers divide the data space in to half-spaces, we know that for any $h \in \mH$, $h^{-1}(-1)$ and $h^{-1}(+1)$ are both convex. Therefore, we get the following corollary.

	\begin{cor}\label{cor:svm_and_relu}Suppose $S$ is linearly separable. Then there exists $\vecv \in \real^d$ such that for all $h \in \mR'_S \cup \mH_S$ and $\vecx\in h^{-1}(-1)$, $\vecv$ is an adversarial direction for $h$ at $\vecx$.\end{cor}

\begin{figure}[ht]
		\centering
		\includegraphics[width=0.5\linewidth]{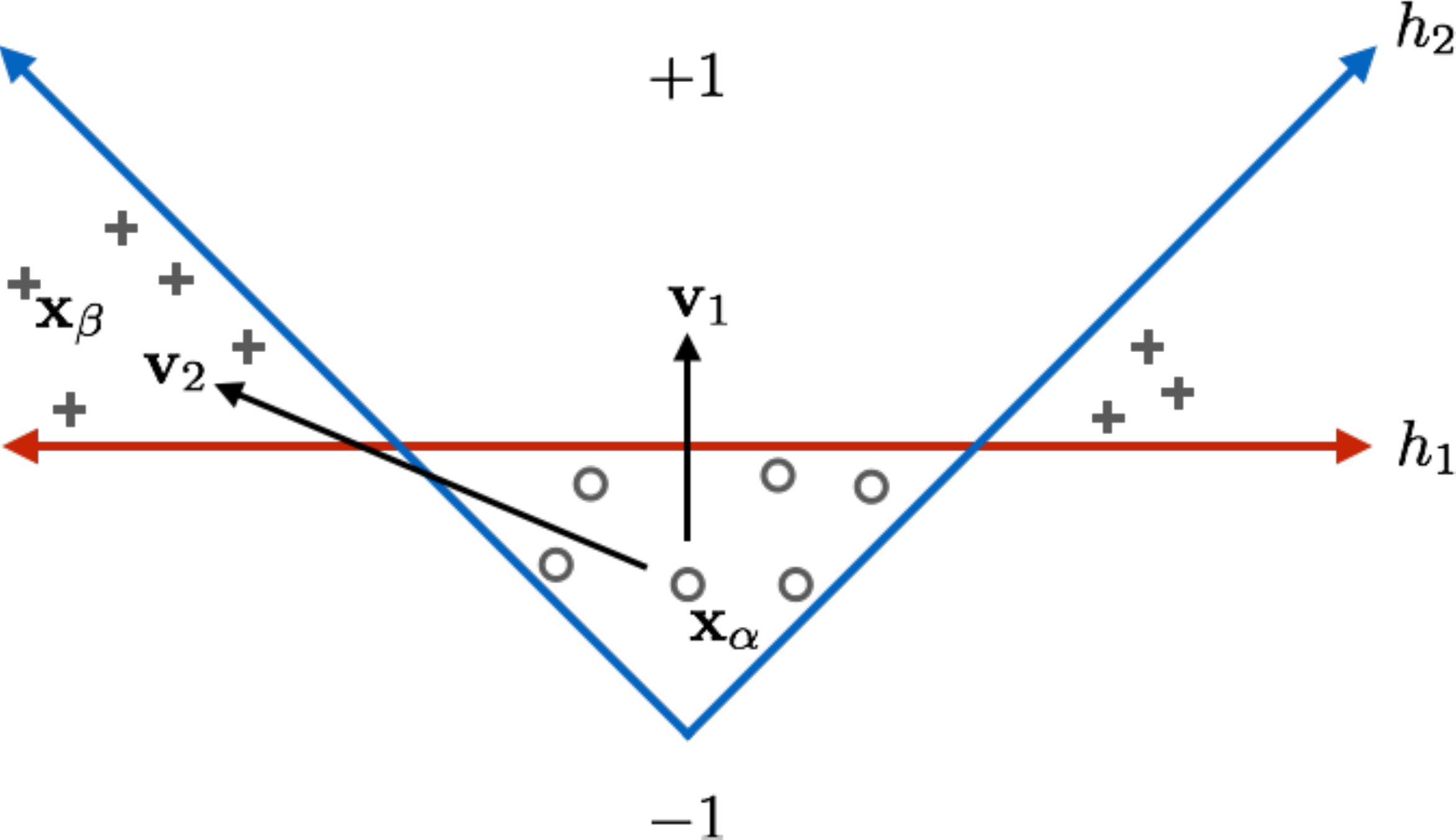}
		\caption{\small The decision boundary of a linear classifier $h_1$ and a two-layer ReLU classifier $h_2$, in red and blue, respectively. The instances with label $1$ are marked by pluses, those with label $-1$ are marked by circles. The vector $\vecv_1$ is the adversarial direction for $h_1$ from Theorem \ref{thm:svm_adv_dir}, and $\vecv_2$ is the ReLU adversarial direction Theorem \ref{thm:relu_direction} pointing from $\vecx_\alpha$ to $\vecx_\beta$.}\label{fig:transfer}
	\end{figure}

	The adversarial direction for $\mH_S$ constructed in Theorem \ref{thm:svm_adv_dir} does not necessarily transfer to $\mR'_S$. A graphical explanation of this is given in Figure \ref{fig:transfer}. As stated in Corollary \ref{cor:svm_and_relu}, the ReLU adversarial direction $\vecv_2$ constructed in the proof of Theorem \ref{thm:relu_direction} transfers to the linear classifier classifier $h_1$. However, the linear classifier adversarial direction $\vecv_1$ constructed in the proof of Theorem \ref{thm:svm_adv_dir} does not serve as an adversarial direction for the $-1$ labeled points for $h_2$. The vector $\vecv_2$ is constructed in the proof of Theorem \ref{thm:convex_adv} in Appendix \ref{append:convex}, and can be any vector $\vecv_2$ in the direction of $\vecx_\beta-\vecx_\alpha$ for any $\vecx_\beta \in S[1], \vecx_\alpha \in S[-1]$.

	This intuitive idea formalized in Appendix \ref{append:svms_relus}. There, we explicitly construct a linearly separable dataset and a ReLU network that correctly classifies the data but where the max-margin classifier does not transfer to the ReLU classifier.


\section{Experiments}\label{sec:experiments}

In this section we empirically study how well adversarial perturbations transfer between classifiers. As demonstrated by \citet{moosavi2017universal}, \citet{narodytska2016simple}, and \citet{papernot2016transferability}, adversarial examples often transfer to other classifiers. Here, we show empirically that adversarial perturbations for ReLU networks often transfer to other ReLU networks and SVMs. Similarly, adversarial perturbations for SVMs trained on different data sets often transfer among one another. Our experiments indicate that adversarial examples generated from SVMs transfer less often to ReLU networks.

\paragraph{Setup: }We use the MNIST dataset to perform our experiments. MNIST consists of handwritten numbers represented as vectors $\vecx \in \real^{28\times 28}$ with labels in $\{0,1,\ldots, 9\}$. We use Pytorch \citep{paszke2017automatic} to train 50 moderately-sized neural networks with the same architecture as the PyTorch MNIST examples. Each neural network is trained on 10000 random images from the MNIST training set. We use scikit-Learn LinearSVC \citep{pedregosa2011scikit,fan2008liblinear} to train 50 support vector machines, each on 10000 random images from the MNIST training set.

To generate adversarial examples, we iterate the Fast Gradient Sign Method (FGSM) of \citet{goodfellow2014explaining}. Define $\mB(\vecx,\epsilon) = \{\vecx' : \|\vecx-\vecx'\|_\infty \leq \epsilon\}$. To find an adversarial $\vecx' \in \mB(\vecx,\epsilon)$, we use projected gradient ascent with step-size $\gamma$ to maximize the loss function $J$. Thus, we move our iterates $\vecx^{(t)}$ in the direction of $\gamma \nabla_{\vecx^{(t)}}(J)$ and then project on to $\mB(\vecx,\epsilon)$.

We perform 40 iterations with $\gamma=.01$ to produce adversarial examples in $\mB(\vecx,\epsilon)$ for $\epsilon$ in $\{0.01, 0.02, \ldots, 0.40\}$. For each classifier and $\epsilon$, we construct 1000 adversarial examples by applying FGSM to 1000 randomly selected examples from the MNIST test set. For each $\epsilon$, this gives us 50000 SVM adversarial examples 50000 neural network adversarial examples. We also generate 50000 random perturbations for a given $\epsilon$.

\begin{figure}[h!]
    \centering
    \begin{subfigure}[t]{0.48\textwidth}
        \centering
        \includegraphics[width=0.9\textwidth]{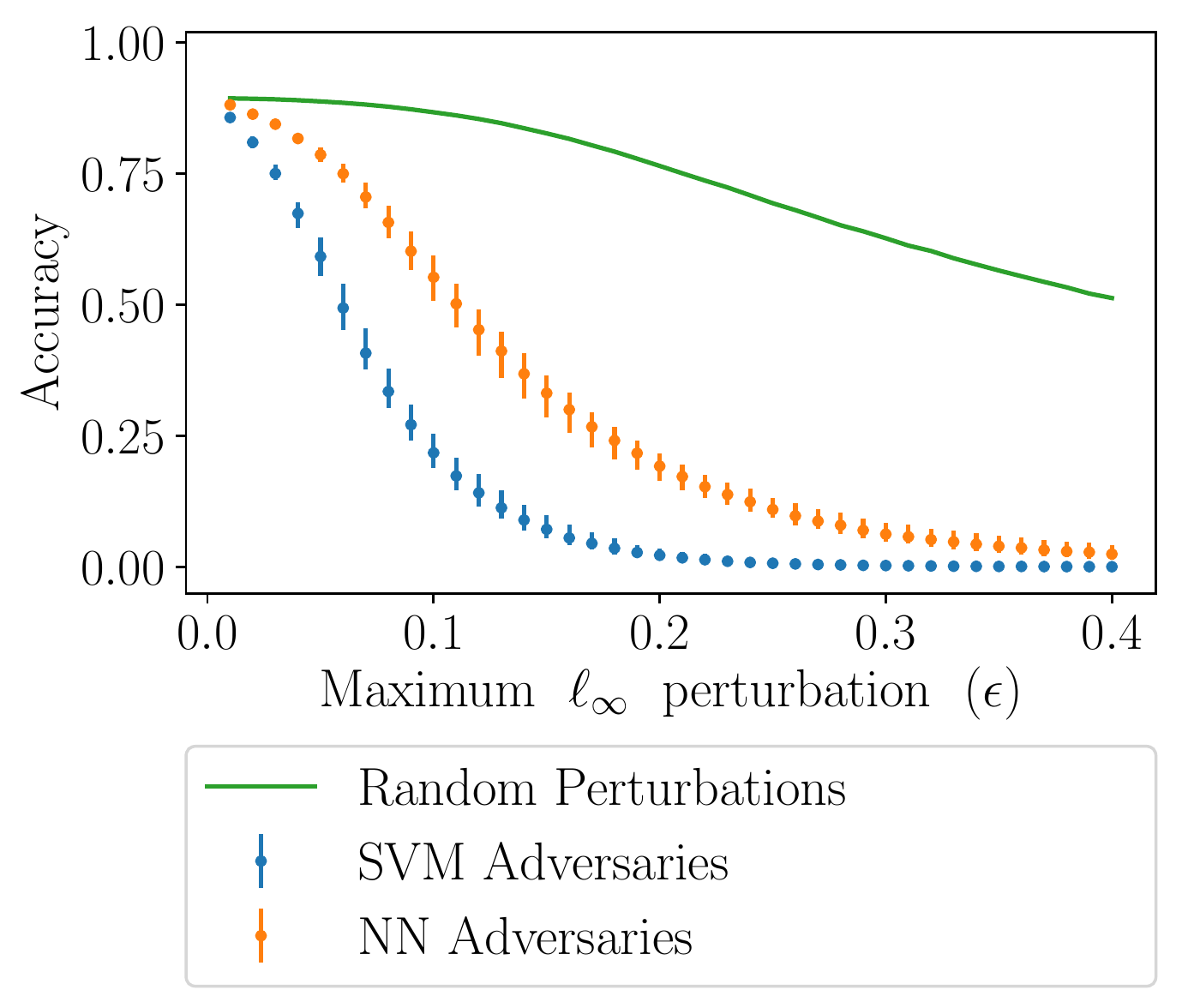}
        \caption{Average accuracy over all trained SVMs on various adversarial examples.}
        \label{subfig:SVM}
    \end{subfigure}\hfill
    \begin{subfigure}[t]{0.48\textwidth}
        \centering
        \includegraphics[width=0.9\textwidth]{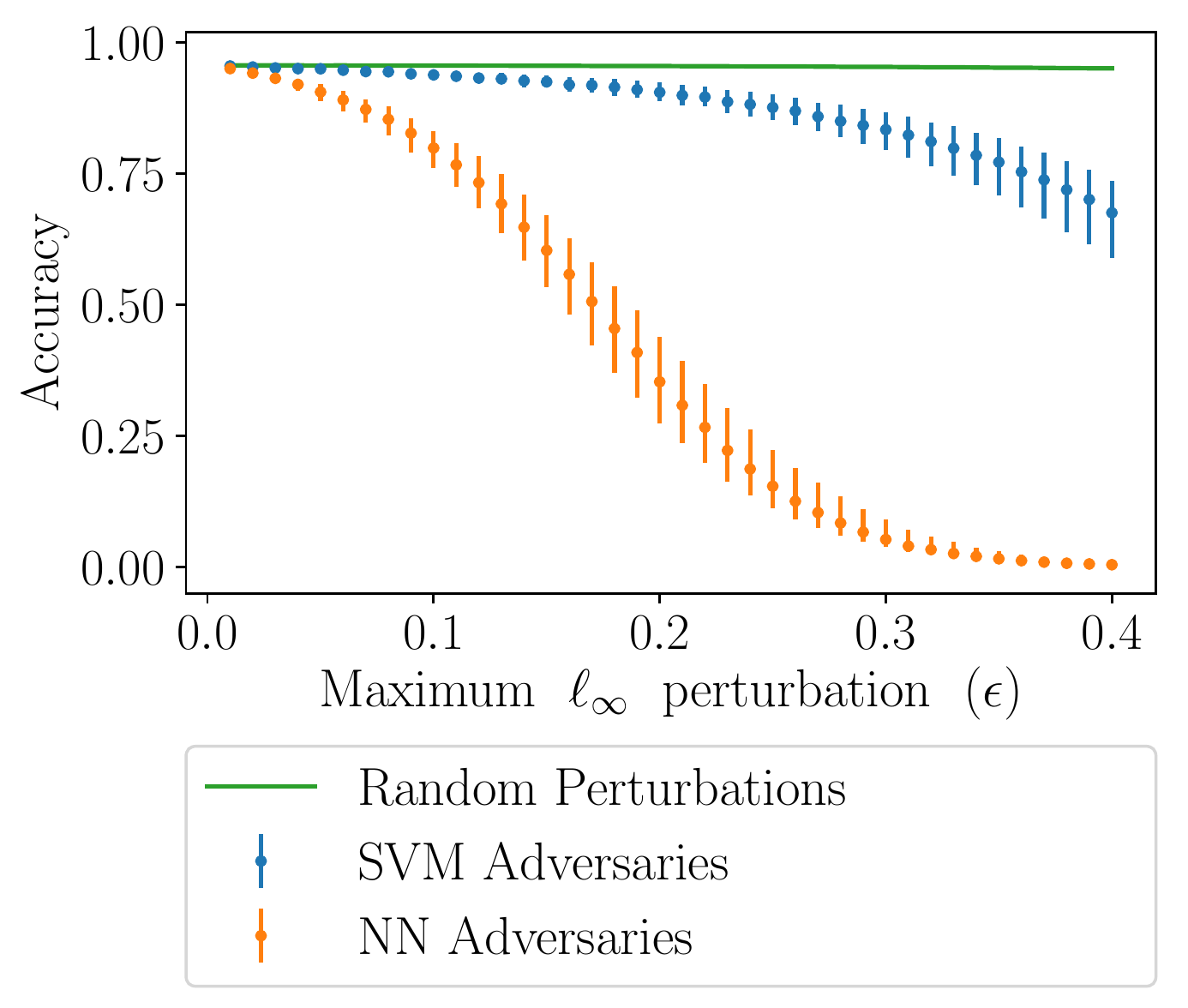}
        \caption{Average accuracy over all trained neural networks on various adversarial examples.}
        \label{subfig:NN}
    \end{subfigure}
    \caption{Average classifier accuracy after applying perturbations of varying $\ell_\infty$ norm. Perturbations are generated randomly, by SVM adversaries, and by neural network (NN) adversaries.}
    \label{fig:MergedPlots}
\end{figure}

For each classifier, we find its overall accuracy on the random perturbations, SVM adversarial examples, and neural network adversarial examples. We plot the average accuracy (over all classifiers of a given type) in Figure \ref{fig:MergedPlots}. We also plot error bars corresponding to the minimum and maximum total accuracy of any single classifier on each type of adversarial examples.

As previously demonstrated in other works (see work by \citet{liu2016delving} and \citet{fawzi2016robustness}), both SVMs and neural networks are relatively robust to random perturbations. Consider Figure \ref{subfig:SVM}. This shows that SVMs are often susceptible to adversarial examples generated from other SVMs trained on different subsets of the data. This supports our result that SVMs trained independently from some distribution are jointly susceptible to some adversarial direction. Moreover, SVMs are still somewhat susceptible to adversarial examples generated from neural networks. While SVM adversaries are more effective than NN adversaries at fooling other SVMs, the neural network adversarial examples still do much better than random perturbations.

Similarly, neural network adversarial examples transfer to other neural networks, as shown in Figure \ref{subfig:NN}. Despite being trained on different subsets of the data, these neural networks are often susceptible to similar perturbations. This stands in stark contrast to the resilience of neural networks to random perturbations. This figure also supports our theory from Section \ref{sec:svms_relus} stating that adversarial examples generated from SVMs need not transfer to neural networks. SVM adversaries only do slightly better than random perturbations at fooling neural networks. This aligns with the empirically observed phenomenon that classifiers with more capacity seem to be more resistant to adversarial attacks \citep{goodfellow2014explaining}.

\section{Conclusion}

In this paper, we theoretically investigate the transferability of adversarial directions on linear classifiers and two-layer ReLU networks. Our results show that classifiers that accurately classify similar datasets are often jointly susceptible to some adversarial direction. This holds even without assumptions on the training procedure. Moreover, we show that while adversarial perturbations for ReLU networks transfer to linear classifiers, the reverse need not hold. We confirm all of our findings experimentally on MNIST.

Adversarial examples and transferability are still not fully understood. While we focus on adversarial directions, a natural extension would be to constrain the norm of the perturbation allowed. Second, our results are derived only for linear classifiers, and could be studied from a kernel perspective as well. Finally, our results above show that ReLU adversarial examples often transfer to linear classifiers. Is this phenomenon more general? Do adversarial examples classifiers with higher effective capacity tend to transfer to those with lower effective capacity? Answers to these questions could help inform future work in designing adversary-resistant classifiers.





\bibliographystyle{plainnat}
\bibliography{transfer_refs}

\newpage
\appendix
\section{Proof of Results}\label{sec:appendix1}

	\subsection{Proof of Theorem \ref{thm:svm_adv_dir}}

		\begin{proof}
			Fix $h \in \mH_S$ where $h(\vecx) = \sgn(\langle \vecw,\vecx\rangle+b)$. Pick $(\vecx,y) \in S$, and suppose $h(\vecx) = y$. We wish to show that for some $c > 0$, $h(\vecx-cy\vecw^*) = -y$. By Lemma \ref{lem:max_marg_w}, we know $\vecw^* = \sum_{i=1}^N \alpha_iy_i\vecx_i$ such that $\sum_{i=1}^N\alpha_iy_i = 0$. Therefore,
			\begin{equation}\label{eq:max_marg_comp}
			\langle \vecw, \vecw^*\rangle = \langle \vecw, \sum_{i=1}^N \alpha_iy_i\vecx_i\rangle = \sum_{i=1}^N \alpha_iy_i\langle \vecw,\vecx_i\rangle.\end{equation}
			We know that $h \in \mH_S$, so $y_i(\langle \vecw,\vecx_i\rangle+b) > 0$ for $1 \leq i \leq N$. Therefore, $y_i\langle \vecw,\vecx_i\rangle > -y_ib$. Plugging this in to \eqref{eq:max_marg_comp} we get
			\begin{equation}\label{eq:0_bound}
			\langle \vecw, \vecw^*\rangle = \sum_{i=1}^N \alpha_iy_i\langle \vecw,\vecx_i\rangle > \sum_{i=1}^N \alpha_i(-y_ib) = 0\end{equation}
			Here we used the fact that $\alpha_i \geq 0$ and $\sum_{i=1}^N \alpha_iy_i = 0$. This then implies
			\begin{equation}
			\langle \vecw, \vecx-cy\vecw^*\rangle +b = \left(\langle \vecw,\vecx\rangle + b\right) - cy\langle \vecw,\vecw^*\rangle.\end{equation}
			Since $\sgn(\langle \vecw,\vecx \rangle + b) = y$ and $\langle \vecw,\vecw^*\rangle > 0$, we know that $\sgn(y\langle \vecw,\vecw^*\rangle ) \neq \sgn(\langle \vecw,\vecx\rangle + b)$. Therefore, by taking $c$ large enough we will have $\sgn(\langle \vecw,\vecx-cy\vecw^*\rangle + b) = -y$.
		\end{proof}	

	\subsection{Proof of Theorem \ref{thm:svm_adv_dir2}}

		\begin{proof}
			The proof is nearly identical to the proof of Theorem~\ref{thm:svm_adv_dir}, except  we use the fact that for any $i$, $y(\langle \vecw,\vecx_i\rangle + b) \geq \gamma_h$. Then \eqref{eq:0_bound} can be refined to
			\begin{equation}\label{eq:gamma_bound}
			\begin{aligned}
			& \langle \vecw, \vecw^*\rangle\\
			&= \sum_{i=1}^N \alpha_iy_i\langle \vecw,\vecx_i\rangle \geq \sum_{i=1}^N \alpha_i(\gamma_h-y_ib) = \gamma_h\sum_{i=1}^N \alpha_i.\end{aligned}
			\end{equation}
			For $c > \dfrac{y(\langle\vecw,\vecx\rangle+b)}{\gamma_h\sum_{i=1}^N \alpha_i}$ we have
			\begin{align*}
			& y(\langle \vecw,x-cy\vecw^*\rangle + b)\\
			&= y(\langle \vecw,\vecx\rangle + b)-c\langle \vecw^*,\vecw\rangle\\
			&< y(\langle\vecw,\vecx\rangle+b) - \dfrac{y(\langle\vecw,\vecx\rangle+b)}{\gamma_h\sum_{i=1}^N\alpha_i}\gamma_h\sum_{i=1}^N\alpha_i\\
			&= 0.\end{align*}
			Therefore, $h(\vecx-cy\vecw^*) \neq h(\vecx)$.
		\end{proof}	

	\subsection{Proof of Theorem \ref{thm:soft_margin}}\label{append:soft_margin}

		\begin{proof}
			Recall that we assume that with probability $1$ over $\mD$, $\|\vecx\|_2 \leq R$. Define
			$$\mH := \{ \vecw \in \real^d \mid \|\vecw\|_2 \leq B\}$$
			and let $\ell: \mH \times \mX\times\mY \to \real$ be a loss function of the form
			\begin{equation}\label{eq:loss_form}
			\ell(\vecw,(\vecx,y)) = \phi(\langle\vecw,\vecx\rangle,y).\end{equation}
			Suppose that for all $y$, $\phi(\cdot, y)$ is $L$-Lipschitz and $|\phi(a,y)| \leq c$ for all $a \in [-BR,BR]$. Then Theorem 26.12 of \citet{shalev2014understanding} states that for any $\delta \in (0,1)$, with probability at least $1-\delta$ over an i.i.d. sample of size $n$ drawn from $\mD$,
			$$\forall\vecw\in\mH,~~L_{\mD}(\vecw) \leq L_{S}(\vecw) + \dfrac{2\rho BR}{\sqrt{n}} + c\sqrt{\dfrac{2\ln(2/\delta)}{n}}$$
			where
			$$L_{\mD}(\vecw) := \EE_{(\vecx,y)\sim\mD}[\ell(\vecw,(\vecx,y))]$$
			$$L_{S}(\vecw) := \dfrac{1}{n}\sum_{(\vecx,y) \in S} \ell(\vecw,(\vecx,y)).$$
			We will apply this to the {\it ramp loss} \citep{huang2014ramp}. The ramp function $\psi: \real \to \real$ is defined by
			$$\psi(u) := \begin{cases} \max\{0,1-u\},~~~u \geq 0\\
			1,~~~u < 0\end{cases}.$$
			The loss function is then given by
			$$\ell(\vecw,(\vecx,y)) = \psi(y\langle \vecw,\vecx\rangle).$$
			In other words, the ramp loss agrees with the hinge loss for $y\langle \vecw,\vecx\rangle \geq 0$ and is otherwise $1$. In particular, $\ell$ is of the form in \eqref{eq:loss_form}, is $1$-Lipschitz, and is bounded by $1$. Moreover, it is simple to see that it gives an upper bound for the $0$-$1$ loss
			$$\ell^{0-1}(\vecw,(\vecx,y)) = \II_{(0,\infty)}(y\langle\vecw,\vecx\rangle).$$
			Therefore, with probability at least $1-\delta$ over $S$, we have
			\begin{align*}
			\forall\vecw\in\mH, &~~\PP_{(\vecx,y)\sim\mD}[y\langle \vecw,\vecx\rangle > 0] \leq L_{\mD}(\vecw)\\
			&\leq L_{S}(\vecw) + \dfrac{2 BR}{\sqrt{n}} + \sqrt{\dfrac{2\ln(2/\delta)}{n}}.\end{align*}
			Let $S^{(j)}, \vecw^{(j)}$ for $j \in \{1,2\}$ be as above. Note that by \textbf{\ref{asm:A1}}, $\vecw^{(j)} \in \mH$. Moreover, by \textbf{\ref{asm:A3}}, $\ell(\vecw^{(j)},(\vecx,y)) = 0$ for at least $(1-\rho)n$ samples in $S^{(j)}$. Since $|\ell(\vecw,(\vecx,y))| \leq 1$, the remaining samples contribute at most $\rho$ to $L_{S^{(j)}}(\vecw^{(j)})$. Therefore, with probability at least $1-2\delta$ over $S^{(1)}$ and $S^{(2)}$, we have that for $j \in \{1,2\}$,
			\begin{equation}\label{eq:prob_wj}
			\PP_{(\vecx,y)\sim\mD}[h_j(\vecx) \neq y] \leq \rho + \dfrac{2 BR}{\sqrt{n}} + \sqrt{\dfrac{2\ln(2/\delta)}{n}}.\end{equation}
			Assume that this holds. By \textbf{\ref{asm:A2}}, we know that for each $(\vecx,y) \in S^{(2)}$, the probability that $h_1$ misclassifies this point is bounded above by \eqref{eq:prob_wj}. Let $X_i$ denote the event that $h_1$ correctly classifies the $i$-th sample of $S^{(2)}$. Thus,
			$$\EE\left[\sum_{i=1}^n X_i\right] \geq (1-\rho)n - 2BR\sqrt{n} - \sqrt{2\ln(2/\delta)n}.$$
			Let $S^{(2)}_1 \subseteq S^{(2)}$ denote the set of points $h_1$ correctly classifies. By Hoeffding's inequality, we find that with probability at least $1-n^{-2}$,
			\begin{align*}
			& |S^{(2)}_1| = \sum_{i=1}^n X_i\\
			&\geq (1-\rho)n - 2BR\sqrt{n} - \sqrt{2\ln(2/\delta)n} - \sqrt{n\ln(n)}.\end{align*}
			Taking $\delta = n^{-2}$, we know that with probability at least $1-n^{-2}$,
			$$|S^{(2)}_1| \geq (1-\rho)n - 2BR\sqrt{n}-4\sqrt{n\ln(n)}.$$
			Let $S^{(2)}_{12} \subseteq S^{(2)}_1 \subseteq S^{(2)}$ be the subset of points that $h_1$ and $h_2$ correctly classify. Since $h_2$ misclassifies at most $\rho n$ points from $S^{(2)}$, we know that with probability at least $1-n^{-2}$,
			\begin{equation}\label{eq:size_bound2}
			|S^{(2)}_{12}| \geq (1-2\rho)n -2BR\sqrt{n}-4\sqrt{n\ln(n)}.\end{equation}
			By the same argument, with probability at least $1-n^{-2}$, the set $S^{(1)}_{12} \subseteq S^{(1)}$ of points correctly classified by $h_1$ and $h_2$ satisfies
			\begin{equation}\label{eq:size_bound1}
			|S^{(1)}_{12}| \geq (1-2\rho)n -2BR\sqrt{n}-4\sqrt{n\ln(n)}.\end{equation}
			Let $S = S^{(1)}_{12}\cup S^{(2)}_{12}$. Taking a union bound with \eqref{eq:prob_wj}, \eqref{eq:size_bound2}, and \eqref{eq:size_bound1} with the value $\delta = n^{-2}$, we find that condition 1 in Theorem \ref{thm:soft_margin} holds and that both $h_j$ are $S$-accurate. We now need to construct an adversarial example. By construction, however, $h_1, h_2$ both linearly separate $S$. Let $\mH_S$ denote the set of linear classifiers that linearly separate $S$. Then by Theorem \ref{thm:svm_adv_dir}, the max-margin classifier $\vecv$ with respect to $S$ gives an adversarial direction for all $h \in \mH_S$ and $(\vecx,y) \in S$, proving condition 2.
		\end{proof}

	\subsection{Proof of Theorem \ref{thm:multi_hard_svm}}

		\begin{proof}
			We will proceed by reducing to the case of single-class linear classification. Initially, one may think that we can simply consider the $k$th linear classifier of $h$. However, as noted above, $S$-accuracy is a weak enough condition that there may be $(\vecx,y) \in S$ such that $y \neq k$ but $(W\vecx+\vecb)_k > 0$. In other words, $S$-accuracy does not require each of the $k$ linear classifiers to be accurate with respect to one-versus-rest classification.

			Instead, select some $l \neq k$. Then for all $(\vecx,k) \in S[k]$, we know that $(W\vecx+\vecb)_k > (W\vecx+\vecb)_l$, while for all $(\vecx,l) \in S[l]$, the reverse holds. Define $\vecw := \vecw_k-\vecw_l$ and $a := b_k-b_l$. Then for all $(\vecx,k) \in S[k]$, $\langle\vecw,\vecx\rangle + a > 0$ while for all $(\vecx,l) \in S[l], \langle\vecw,\vecx\rangle + a < 0$. Therefore, the linear classifier $h$ associated to $(\vecw,a)$ is $S[k]\cup S[l]$ accurate. Hence, there is some max-margin classifier $h^*$ on this instance set with associated tuple $(\vecw^*,a^*)$. In particular, by Theorem \ref{thm:svm_adv_dir} we know that $\vecv = -\vecw^*$ is an adversarial direction for $h$ at all $(\vecx,y) \in S[k]$.
		\end{proof}

	\subsection{Proof of Theorem \ref{thm:multi_soft_svm}}

		\begin{proof}
			We will apply similar techniques to the proof of Theorem \ref{thm:multi_hard_svm} in order to reduce this to a situation similar to that of Theorem \ref{thm:soft_margin}. Select $l$ as in \textbf{\ref{asm:B3}} and fix $j \in \{1,2\}$. Define $\vecw^{(j)} := \vecw^{(j)}_k-\vecw^{(j)}_l$. Let $T^{(j)}$ denote the set $S^{(j)}[k] \cup S^{(j)}[l]$, but with the label $k$ replaced by $1$ and the label $l$ replaced by $-1$.

			By the triangle inequality and \textbf{\ref{asm:B1}}, we know that $\|\vecw^{(j)}\|_2 \leq 2B$, and that \textbf{\ref{asm:B2}} implies that whether $\vecw^{(1)}$ correctly classifies some point in $T^{(1)}$ is independent from $(T^{(2)},\vecw^{(2)})$ and vice-versa. By \textbf{\ref{asm:B3}}, we know that there are at least $(1-\rho)n_{j,k}$ points $(\vecx,1) \in T^{(j)}[k]$ such that $\langle \vecw^{(j)},\vecx\rangle > 1$.

			Let $\mD[k,l]$ be the marginal distribution of $\mD$ on $\mX \times \{k,l\}$, and let $\mD'$ denote the corresponding distribution on $\mX \times \{\pm 1\}$ given by replacing $k$ by $1$ and $l$ by $-1$. Note that $T^{(1)}$ and $T^{(2)}$ are both sets of size $n$ drawn i.i.d. from the distribution $\mD'$.

			We can then apply Theorem \ref{thm:soft_margin} to $T^{(1)}$ and $T^{(2)}$. This implies that there is a set $T \subseteq T^{(1)} \cup T^{(2)}$ and a perturbation $\vecv$ such that $|T\cap T^{(j)}| \geq (1-2\rho)n - 4BR - 4\sqrt{n\ln(n)}$ and for all $(\vecx,y) \in T$, $-y\vecv$ is an adversarial direction for $\vecw^{(j)}$ at $\vecx$. Note that we obtain a $4BR$ term instead of a $2BR$ term due to the fact that $\|\vecw^{(j)}\|_2 \leq 2B$ instead of $B$.

			We define $S$ to be $T$ but replacing the labels $1, -1$ with $k,l$. Therefore, $S$ satisfies condition 1 of the theorem. It now suffices to show condition 2. Suppose $(\vecx,y) \in S\cap S^{(j)}[k]$ and $h_j(\vecx) = k$. Then $\langle \vecw^{(j)},\vecx\rangle > 0$. By construction of $\vecv$, $-\vecv$ is an adversarial direction for $\vecw^{(j)}$, so there is some $c$ such that $\langle \vecw^{(j)},\vecx-c\vecv\rangle < 0$. This implies that $(W^{(j)}\vecx)_l > (W^{(j)}\vecx)_k$, so $h_j(\vecx) \neq k$. We can argue analogously for $(\vecx,y) \in S\cap S^{(j)}[l]$, completing the proof.
		\end{proof}

	\subsection{Proof of Theorem \ref{thm:convex_adv}}\label{append:convex}

		\begin{proof}
			Fix $h \in \mC_S$ and let $A = \overline{h^{-1}(\alpha)}$. Then $A$ is a closed convex set. Therefore, $A$ is the intersection of affine hyperplanes. That is, there is some collection of hyperplanes indexed by a set $I$ where
			$$A = \bigcap_{i \in I} \left\{\vecx \in \real^d ~\middle|~ \vecw_i^T\vecx + b_i \leq 0\right\}.$$
			Select some $\vecx_\alpha \in S[\alpha]$ and $\vecx_\beta \in S[\beta]$, and let $\vecv = \vecx_\beta-\vecx_\alpha$. Note that we then have for all $i \in I$,
			$$\vecw_i^T\vecv = \vecw_i^T\vecx_\beta - \vecw_i^T\vecx_\alpha \geq b_i-b_i = 0.$$
			Note however that since $\vecx_\alpha, \vecx_\beta$ are assigned different labels by $h$ (as $h$ is $S$-accurate), it cannot be the case that $\vecw_i^T\vecx_\alpha + b_i = 0 = \vecw_i^T\vecx_\beta + b_i$. Therefore, $\vecw_i^T\vecv > 0$. Hence, for any $\vecx \in h^{-1}(\alpha)$, there is some $c > 0$ such that for all $i \in I$,
			$$\vecw_i^T(\vecx + c\vecv) + b_i = \vecw_i^T\vecx+b_i + c \vecw_i^T\vecv > 0.$$
			In particular, this implies that $\vecx+c\vecv \notin A$, and so $h(\vecx+c\vecv) \neq h(\vecx)$. Since the choice of $\vecv$ was independent of $h$, we find that $\vecv$ is adversarial for all $h \in \mC_S$.
		\end{proof}	

	\subsection{Proof of Lemma \ref{lem:convex_class}}

		\begin{proof}
			By definition of $\mR$, we can write $h(\vecx) = g(f(\vecx))$ where $f(\vecx) = \sum_{i=1}^L \vecw_i^T\vecx+b_i$ and $g(z) = \JJ_{(0,\infty)}(z)$. Therefore, $h(\vecx) = 1$ iff $\vecw_i^T\vecx+b_i > 0$ for some $i$. In particular, this implies
			$$h^{-1}(-1) = \bigcap_{i=1}^L \left\{ \vecx\in\real^d ~\middle|~ \vecw_i^T\vecx+b_i \leq 0\right\}.$$
			This is a convex region.
		\end{proof}

	\subsection{Proof of Lemma \ref{lem:mr1_conv}}

		\begin{proof}
			Suppose $h(\vecx) = g(f(\vecx))$ where $f(\vecx) = \vecv^T\sigma(W\vecx+\vecb)$, $\vecv \geq 0$, and $g(z) = \mathbb{J}_{[a,\infty)}(z)$. The case where $g(z) = \mathbb{J}_{(a,\infty)}(z)$ follows from a virtually identical argument. Let $\vecw_i^T$ denote the $i$th row of $W$. Let $A = h^{-1}(1)$. We also define a set $B$ as follows:
			$$B := \bigcup_{\substack{I \in \mP([L])\\I \neq \emptyset}} B_I.$$
			$$B_I := \left\{ \vecx\in \real^d ~\middle|~ \left\langle \sum_{i \in I} v_i\vecw_i,~\vecx\right\rangle + \sum_{i \in I} v_ib_i \geq a\right\}.$$

			Here, $\mP([L])$ denotes the power set of $\{1,\ldots, L\}$. Note that $B$ is a union of closed half-spaces, so its complement $B^c$ is convex. We wish to show that $A = B$. This then implies that $h^{-1}(-1) = A^c = B^c$ is convex. For a given $i \in [L]$, we will let $f_i(\vecx) = v_i\sigma(\vecw_i^T\vecx+b_i)$. Then note that $f(\vecx) = \sum_{i=1}^L f_i(\vecx)$ and each $f_i$ is nonnegative (as $v_i \geq 0$ for all $i$). Let $f_I(\vecx) = \sum_{i \in I} f_i(\vecx)$. Therefore, for any $I \in \mP([L])$, we have
			$$f(\vecx) = f_I(\vecx) + f_{I^c}(\vecx).$$
			Since $f_{I^c}(\vecx) \geq 0$, we know that $f(\vecx) \geq f_I(\vecx)$ for all $I$. Let $q_i(\vecx) = v_i\vecw_i^T\vecx + v_ib_i$ and $q_I(\vecx) = \sum_{i \in I}q_i(\vecx)$. Then clearly $q_i(\vecx) \leq f_i(\vecx)$ and
			$$B_I = \left\{ \vecx\in\real^d~\middle|~q_I(\vecx) \geq a\right\}.$$
			Therefore, if $\vecx \in B_I$, then $q_I(\vecx) \geq a$ and so $f_I(\vecx) \geq a$, which finally implies $f(\vecx) \geq a$. Therefore $B_I \subseteq A$ for all $I$.

			In the reverse direction, suppose $\vecx \in A$. Let $I$ denote the set of indices $i$ such that $f_i(\vecx) = \sigma(\vecw_i^T\vecx+b_i) > 0$. Then $f(\vecx) = f_I(\vecx)$. However, since each $f_i(\vecx) > 0$, by definition of $\sigma$ this implies that $q_i(\vecx) > 0$. Therefore, $q_I(\vecx) = f_I(\vecx) = f(\vecx) > a$, and so $\vecx \in B_I \subseteq B$.
		\end{proof}	

	\subsection{Linear Classifier Adversarial Directions Need Not Transfer to ReLUs}\label{append:svms_relus}

		In this section, we explicitly construct a linearly separable dataset and a ReLU network that correctly classifies the data but where the max-margin classifier does not transfer to the ReLU. 

		Consider the training set $S$ with three points $(\vecx_i,y_i)$ where
		\begin{align*}
		\vecx_1 = (-1,1), y_1 = 1\\
		\vecx_2 = (1,1), y_2 = 1\\
		\vecx_3 = (0,-1), y_3 = -1.\end{align*}

		It is straightforward to see that the max-margin classifier $\vecw^*, b^*$ is given by $\vecw^* = (0,1)$, $b^* = 0$. Therefore, the adversarial direction specified by Theorem \ref{thm:svm_adv_dir} is $\vecw^* = (0,1)$. However, we could consider the following ReLU network:
		$$f(\vecx) = \vecv^T\sigma(W\vecx + \vecb)$$
		where
		$$W = \begin{pmatrix}1 & 1/4\\
		-1 & -1/4\end{pmatrix},~\vecb = \begin{pmatrix}1/2\\ -1/2\end{pmatrix}~,\vecv = \begin{pmatrix}1\\1\end{pmatrix}.$$

		Here, $\sigma$ denotes the coordinate-wise ReLU function. Our label function is given by $g(f(\vecx))$ where $g(z) = -1$ if $z > 0$ and $1$ otherwise. It is clear to see that $g(\vecx_i) = y_i$ for all $i$. Moreover, for all $c > 0$, $g(\vecx_3+c\vecw^*) = -1$. Therefore, $\vecw^*$ is not an adversarial direction for this ReLU network at $(\vecx_3,y_3)$.

\end{document}